%% file: main.tex
\documentclass{article}

\usepackage[final]{neurips_glfrontiers_2023}

\usepackage[utf8]{inputenc} 
\usepackage{amsmath}
\usepackage[T1]{fontenc}    
\usepackage{hyperref}       
\usepackage{url}            
\usepackage{booktabs}       
\usepackage{amsfonts}       
\usepackage{nicefrac}       
\usepackage{microtype}      
\usepackage{xcolor}         
\usepackage{multirow}
\usepackage{wrapfig}
\usepackage{algorithm}
\usepackage{algpseudocode}
\usepackage{graphicx}

\newcommand{\calG}{\mathcal{G}}

\newcommand{\bfA}{\mathbf{A}}
\newcommand{\bfP}{\mathbf{P}}
\newcommand{\bfX}{\mathbf{X}}
\newcommand{\Ltrain}{\mathcal{L}_{\text{train}}}
\newcommand{\Latk}{\mathcal{L}_{\text{atk}}}
\newcommand{\Lpois}{\mathcal{L}_{\text{pois}}}
\newcommand{\Lev}{\mathcal{L}_{\text{ev}}}

\DeclareMathOperator*{\argmax}{argmax}
\DeclareMathOperator*{\argmin}{argmin}

\title{Poisoning \(\times\) Evasion: Symbiotic Adversarial Robustness for Graph Neural Networks}

\author{
  Ege Erdogan, Simon Geisler, Stephan Günnemann \\
  \texttt{\{ege.erdogan, s.geisler, s.guennemann\}@tum.de} \\
  Department of Computer Science \\
  Technical University of Munich 
}

\begin{document}

\maketitle

\begin{abstract}
It is well-known that deep learning models are vulnerable w.r.t.\ small input perturbations. Such perturbed instances are called adversarial examples.  Adversarial examples are commonly crafted to fool a model either at training time (poisoning) or test time (evasion). In this work, we study the symbiosis of poisoning and evasion. We show that combining both threat models can substantially improve the devastating efficacy of adversarial attacks. Specifically, we study the robustness of Graph Neural Networks (GNNs) under structure perturbations and devise a memory-efficient adaptive end-to-end attack for the novel threat model using first-order optimization.

\end{abstract}

\section{Introduction}

\begin{wrapfigure}{r}{0.45\textwidth}
    \centering
    \includegraphics[width=\linewidth]{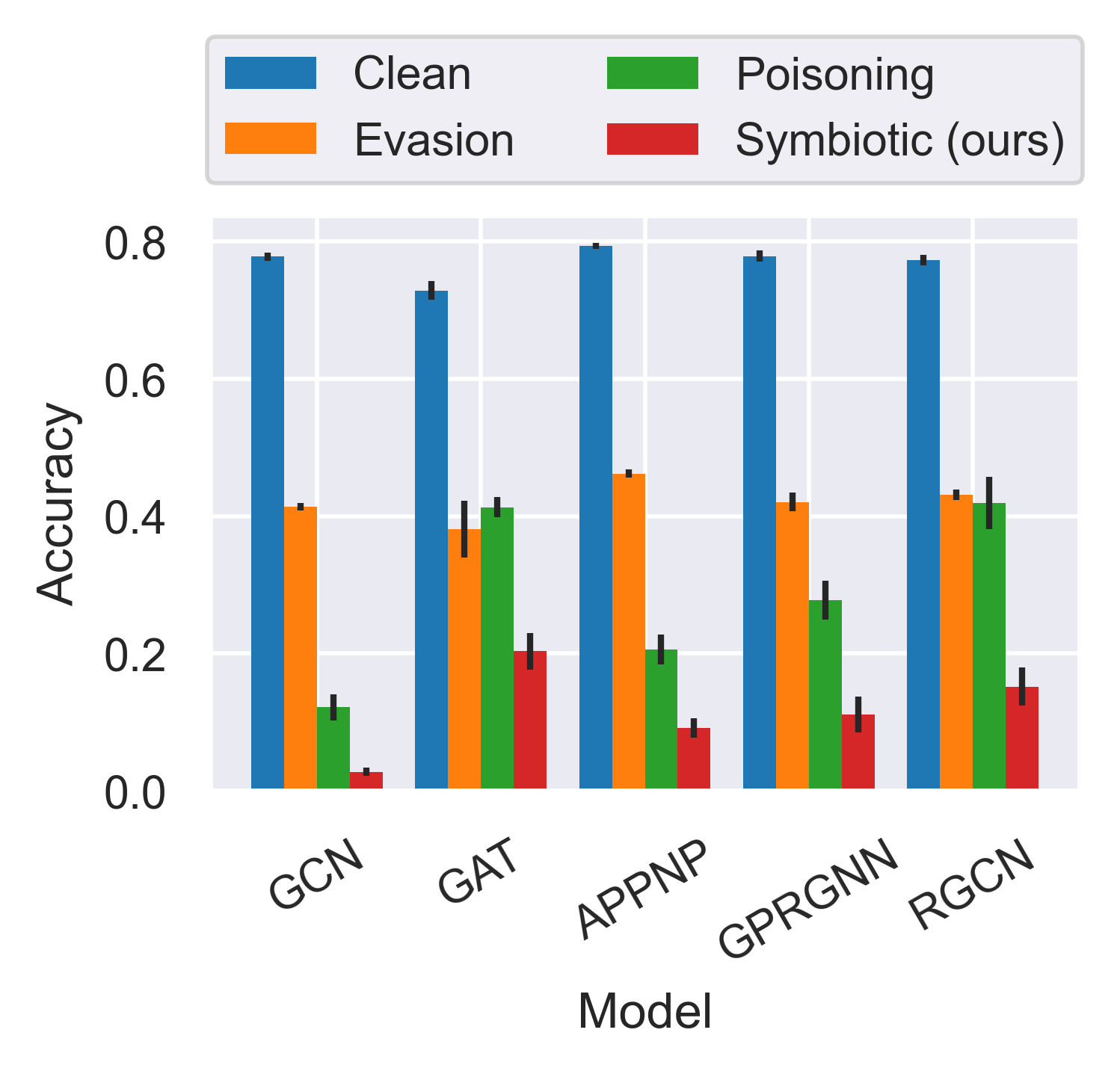}
    \caption{Perturbed accuracies (with standard errors) after evasion, poisoning, and symbiotic (ours) attacks on different models on PubMed.}
    \label{fig:pubmed}
\end{wrapfigure}

Graph neural networks (GNNs) are increasingly used in different domains such as product recommendations \citep{hao2020p} and drug discovery \citep{guo2021dockstream}. Nevertheless, GNNs are vulnerable to adversarial attacks across many tasks such as node classification \citep{zugnerAdversarialAttacksNeural2018, daiAdversarialAttackGraph2018, geislerRobustnessGraphNeural2021}, graph classification \citep{daiAdversarialAttackGraph2018, wang2023revisiting}, link prediction \citep{chen2020link}, and node embeddings \citep{bojchevskiAdversarialAttacksNode2019, zhangDataPoisoningAttack2019}. With attacks being able to scale to very large graphs \citep{geislerRobustnessGraphNeural2021}, studying the adversarial robustness of GNNs is of growing importance. GNNs can be attacked during test time (evasion) or train time (poisoning); yet, a threat model combining evasion and poisoning has not been considered in the literature. It is nevertheless a reasonable threat model considering e.g. publicly available graphs, or graphs extracted from sources such as social media sites.

\textbf{The Problem.} We consider node classification tasks, and an adversary able to change the structure of the graph (i.e. insert/remove edges), with both train and test time access to the graph. The adversary's end goal is to minimize the classification accuracy on the test set. It is constrained by a global budget and attacks entire graph at once rather than targeting specific nodes. We call such attacks combining evasion and poisoning in this threat model \textit{symbiotic} attacks.

\textbf{Contributions.} We initiate the study of this threat model and compare it with plain poisoning and evasion adversaries by adapting the previous PR-BCD attack \citep{geislerRobustnessGraphNeural2021} to the symbiotic threat model, resulting in memory-efficient attacks that can scale to large graphs. Our main findings are:
\begin{itemize}
    \item The symbiotic attacks are consistently stronger than a poisoning attack, indicating that it is beneficial to allocate part of the resources of the poisoning attack for the evasion objective. 
    \item Evasion attacks are constrained by the number of test nodes, with larger test sets making an evasion attack harder. The symbiotic attacks are affected less significantly by the size of the test set since they can also manipulate the graph during training, leading to almost-zero accuracy in cases such as when the share of labeled train is low and test nodes high.  
\end{itemize}

Overall, the significance of the potential improvement the symbiotic threat model provides indicates that it requires further study, and we outline paths of future work in our conclusion.

\section{Preliminaries}

\textbf{Notation.} Throughout we denote a graph by $\calG$ with $n$ nodes, adjacency matrix $\bfA \in \{0,1\}^{n \times n}$ and feature matrix $\bfX \in \mathrm{R}^{n \times d}$, and a GNN applied to the graph as $f_\theta(\calG)$ with parameters $\theta$. $\Phi(\calG)$ is the set of admissible adversarial graphs resulting from $\calG$, and $\Latk$ and $\Ltrain$ denote the adversarial and training objectives. 

\subsection{Adversarial Robustness of GNNs}

An adversarial attack on a GNN can change the structure of the graph by inserting/removing edges and nodes, or can modify the node features. We focus on node classification and edge-level structural perturbations. 

Attacks can be divided into two categories: \textit{evasion} and \textit{poisoning}. An evasion attack targets a fixed GNN with $\theta$ obtained on a clean graph, and thus tries to solve the optimization problem 
\begin{equation}
    \max_{\hat{\calG} \in \Phi(\calG)}
        \Latk(f_\theta(\hat{\calG})),
\end{equation}
while a poisoning attack is performed before training but aims to degrade performance after training:
\begin{equation}
    \max_{\hat{\calG} \in \Phi(\calG)}
        \Latk(f_{\theta^*}(\hat{\calG})) \quad \text{where} \quad
    \theta^* = \argmin_{\theta} \Ltrain(f_\theta(\hat{\calG})).
    \label{eg:poisoning_problem}
\end{equation}
A poisoning attack is admittedly more challenging. Previous work has tried using evasion perturbations as poisoning perturbations \citep{zugnerAdversarialAttacksNeural2018}, or unrolling the training procedure as part of the optimization to compute meta-gradients (gradients w.r.t. hyperparameters) of $\Latk$ w.r.t. $\bfA$ \citep{zugnerAdversarialAttacksGraph2020}.

Finally, since we only consider changes to the binary adjacency matrix, we define $\Phi(\calG)$ to include graphs reachable from $\calG$ after at most $\Delta$ edge perturbations (e.g. for undirected $\calG$, $\Phi(\calG) = \{\tilde{\calG} \,|\, \|\tilde{\bfA} - \bfA\|_0 \le 2 \Delta \,\land\, \tilde{\bfA}^\top = \tilde{\bfA}\}$), though it would also be straightforward to use metrics such as the graph's degree distribution or diameter. 

\subsubsection{PR-BCD}

Our work builds on the \textit{Projected Randomized Block Coordinate Descent} (PR-BCD) attack proposed in \citet{geislerRobustnessGraphNeural2021}. Similar to the Projected Gradient Descent (PGD) attack \citep{xuTopologyAttackDefense2019}, the adjacency matrix is relaxed to $\bfP \in [0,1]^{n \times n}$ to enable continuous gradient updates, and each entry denotes the probability of flipping that edge with the final perturbations sampled from $\text{Bernoulli}(\bfP)$. However, since the adjacency matrix grows quadratically with the number of nodes, scalability of plain PGD becomes a challenge on larger graphs. 

PR-BCD employs Randomized Block Coordinate Descent (R-BCD) \citep{nesterovEfficiencyCoordinateDescent2012} and updates a block of size $b$ of $\bfP$ in each iteration. The projection step ensures that the budget is enforced in expectation; i.e. $\mathrm{E}[\text{Bernoulli}(\bfP)] = \sum \bfP \leq \Delta$ and that $\bfP \in [0,1]^{n \times n}$. After each iteration, rather than sampling the entire block again, the promising entries of the block are kept and only the rest is resampled. In other words, $\bfP$ is kept sparse in a survival-of-the-fittest manner. Multiple samples are drawn at the end and the best-performing one is returned as the final perturbed graph. 

PGD can also be applied for a poisoning attack as the Meta-PGD attack in \citet{mujkanovicAreDefensesGraph2022}. In our attacks, we employ the same principle with PR-BCD to scale better to larger graphs. 

While we only consider a single global budget \(\Delta\), it is straightforward to include more sophisticated constraints when beneficial for the application at hand \citep{gosch_revisiting_2023, gosch_adversarial_training_2023}.

\section{Symbiotic Attacks}

\textbf{The Symbiotic Objective}. The problem of a symbiotic attack has a similar form to the bi-level optimization of Equation \ref{eg:poisoning_problem}, but the main objective is conditioned on the evasion graph $\calG^*$ in addition to the parameters $\theta^*$:
\begin{align}\begin{split}
    \label{eq:joint_problem}
    \max_{\hat{\calG} \in \Phi(\calG)}
        \Lpois(f_{\theta^*} (\calG^*)) \quad &\text{where} \quad
    \theta^* = \argmin_{\theta} \Ltrain(f_\theta(\hat{\calG})), \\ 
        &\text{and} \quad \calG^* = \argmax_{\tilde{\calG} \in \Phi(\hat{\calG})} \Lev(f_{\theta^*}(\tilde{\calG})) 
\end{split}\end{align}
where for clarity we separate the poisoning and evasion objectives $\Lpois$ and $\Lev$ although they might be the same function. 

\textbf{Threat Model.} We model an attacker with the goal of degrading a model's performance on node classification tasks. Combining poisoning and evasion adversaries, our attacker has full access to the graph both at train and test times, knows the model architecture being used so that it can create surrogate models, but can access the trained model only as a black-box. Finally, our attacker is limited by a global budget of edge insertions/removals. 

With the symbiotic objective and the threat model, we propose two attacks as approximations to the optimal solution.

\textbf{The Sequential Attack.} A simple way of launching a symbiotic attack is to split the budget into two, and then to launch an evasion attack with the second part after poisoning the model using the first part. In this case, the poisoning attack has no knowledge of a future evasion, but it can can still help the evasion attack by causing some nodes to be misclassified and reducing the classification margin of some nodes, making them easier to be misclassified during an evasion attack.

\textbf{The Joint Attack}. The poisoning attack can also be designed to ``fit" the future evasion graph by including the evasion attack within the poisoning loss. In this \textit{joint} attack, an evasion attack is launched within each poisoning attack iteration, and the poisoning loss is computed with the poisoned model over the evasion graph, so that the poisoning attack not only reduces the model's accuracy but also makes it more vulnerable to evasion. 

Although the joint and sequential attacks can be instantiated using different evasion/poisoning attacks, we choose to build on PR-BCD since it scales well to larger graphs. This is a significant consideration since especially for the joint attack the inner evasion attack has to performed many times. Also note that the sequential attack is in fact a special case of the joint attack with zero iterations per inner evasion attack.

\section{Evaluation}

\begin{wraptable}{r}{0.45\textwidth}
    \vspace{-1.7cm}
    \centering
    \caption{Numbers of nodes, edges, and classes in the datasets we include in our evaluations. \\}
    \label{tab:num_nodes_edges}
    \begin{tabular}{lccc}
        \toprule
         \textbf{Dataset} & Nodes  & Edges & Classes \\
         \midrule
         Cora       & 2,708  & 10,556 & 7 \\
         CiteSeer   & 3,327  & 9,104  & 6 \\
         PubMed     & 19,717 & 88,648 & 3 \\
         \bottomrule
    \end{tabular}
\end{wraptable}

\subsection{Setup} 

In this section, we compare the symbiotic threat model with evasion and poisoning attacks, instantiated through the PR-BCD evasion \citep{geislerRobustnessGraphNeural2021} and poisoning attacks \citep{mujkanovicAreDefensesGraph2022} on Cora, CiteSeer \citep{mccallum2000automating}, and PubMed \citep{sen2008collective}. We study the robustness of GCN \citep{kipfSemiSupervisedClassificationGraph2016}, GAT \citep{velickovicGraphAttentionNetworks2018a}, APPNP \citep{gasteigerPredictThenPropagate2018}, and GPRGNN \citep{chienAdaptiveUniversalGeneralized2020}. We also consider R-GCN \citep{zhuRobustGraphConvolutional2019} and Jaccard purification \citep{wuAdversarialExamplesGraph2019} as potential defensive mechanisms. For each dataset, we allocate 20 nodes of each class for the labeled training set and allocate 10\% of the nodes as the test set. For transductive tasks, the test nodes are also included as unlabeled train nodes during training, whereas for inductive tasks the test nodes are added to the graph after training. Table \ref{tab:num_nodes_edges} displays the numbers of nodes, edges, and classes in each of our benchmark datasets.

\begin{figure}
    \begin{center}
        \includegraphics[width=0.9\textwidth]{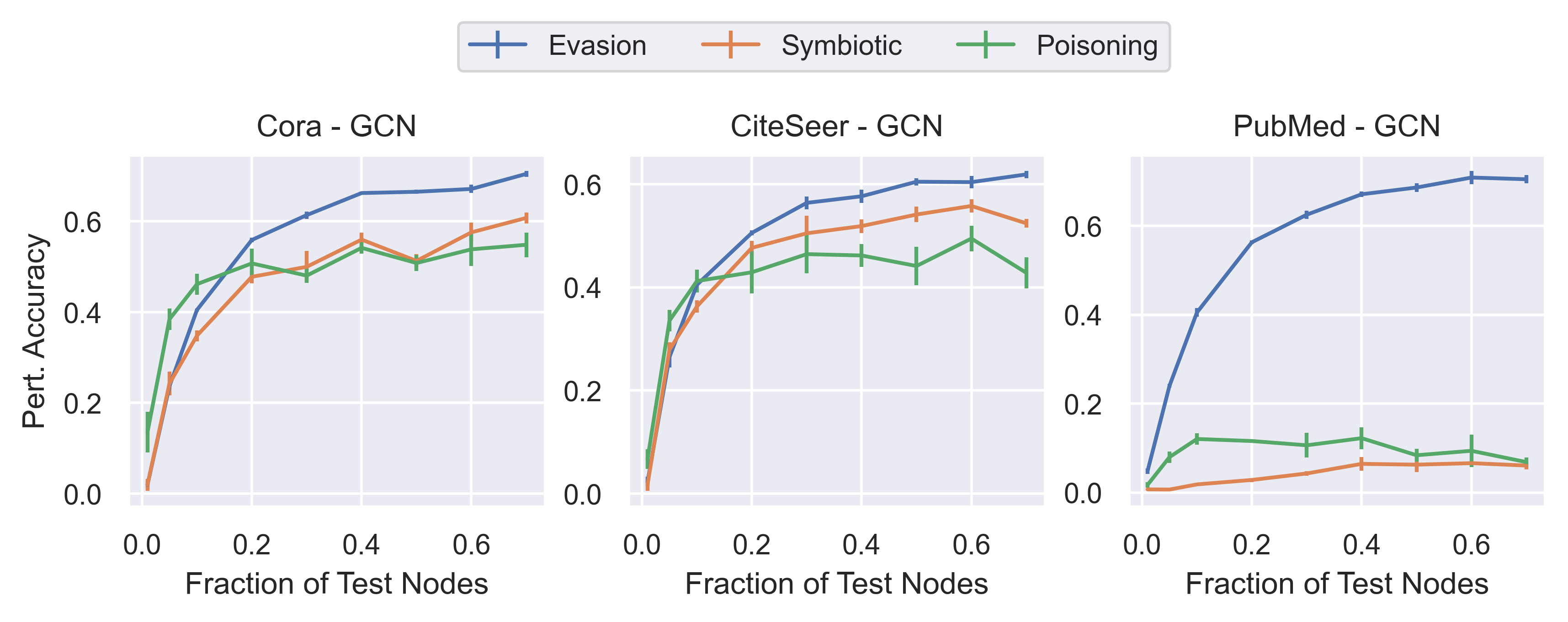}
    \end{center}
    \caption{Perturbed accuracies and standard errors after the four attacks with different test set sizes and a fixed 5\% global budget, using a GCN on the three benchmark datasets.}
    \label{fig:test_size}    
\end{figure}

\subsection{Results} 

Table \ref{tab:main_results} displays the perturbed accuracy values on the test set (10\% of nodes) for our benchmark datasets/models averaged over 10 runs, along with the standard error of the mean. We limit the attacker with a budget 5\% the number of edges, with the budget split equally between poisoning and evasion for the symbiotic attacks. To better focus on the threat model, we report the better-performing of the two attacks for the symbiotic threat model and leave a comparison between the attacks to the appendix. 

Symbiotic attacks are stronger than poisoning across all tasks, and stronger than plain evasion especially against GCN and GPRGNN models. The effect of the symbiotic threat model is most evident on the larger PubMed graph, with the accuracy dropping to almost zero e.g. against a GCN, especially as the evasion attack's performance drops against larger test sets as we discuss next.    

\input{tables/main_results} 

\begin{figure}[t!]
    \centering
    \includegraphics[width=0.9\textwidth]{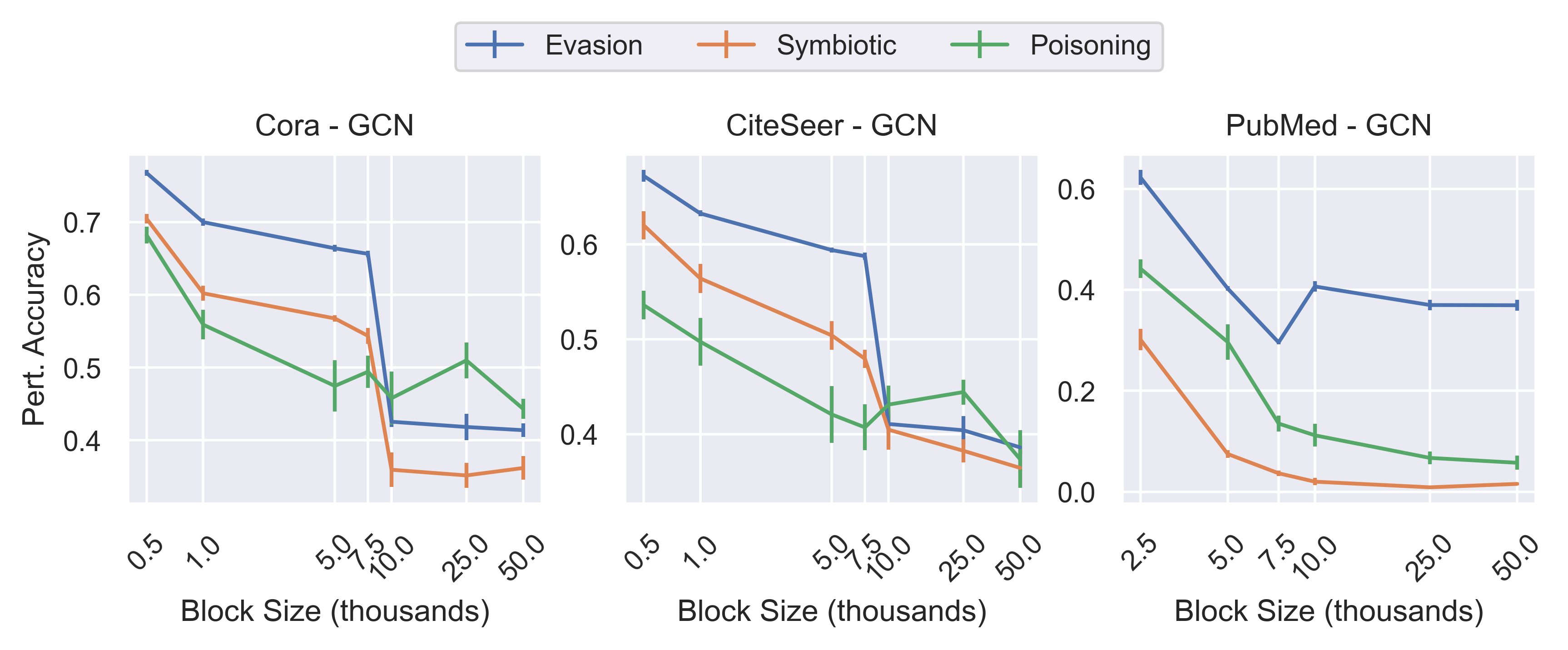}
    \caption{Classification accuracy after the four attacks on a GCN, with the varying \textbf{block sizes} for the PR-BCD optimization displayed along the x-axis.}
    \label{fig:block_size}
\end{figure}

\subsection{Effect of the Number of Test Nodes} 

To illustrate a fundamental point of difference between poisoning and evasion objectives, Figure \ref{fig:test_size} displays the perturbed accuracies for evasion, poisoning, and symbiotic attacks over varying fractions of test nodes using a GCN with a 5\% global budget. Attacking a very small number of nodes is easy as all attacks can obtain zero accuracy since a few edge perturbations should be enough to manipulate a small number of nodes. 

As the number of test nodes grows, evasion becomes considerably more difficult across all datasets. Although poisoning and symbiotic attacks become more difficult as well with more test nodes, especially on PubMed they are much more robust than the evasion attack. Thus the degrading performance cannot be explained only by the attacks having to use the same budget to target a larger number of nodes. The poisoning attack is more lightly affected as it can also manipulate the flow of information across the graph during training; e.g. the PubMed in our setup has the lowest share of labeled train nodes (20 nodes per class), which leads to poisoning attacks having devastating impact. 

Similarly for the symbiotic attacks, poisoning helps in two ways: by reducing the base accuracy evasion starts with, and possibly changing the structure of the graph in a more effective way to block the flow of label information from the labeled train nodes. This results in symbiotic attacks being both more robust against larger test sets than plain evasion, while also being stronger than poisoning alone. 

\begin{figure}[t!]
    \vspace{-0.2cm}
    \centering
    \includegraphics[width=0.9\textwidth]{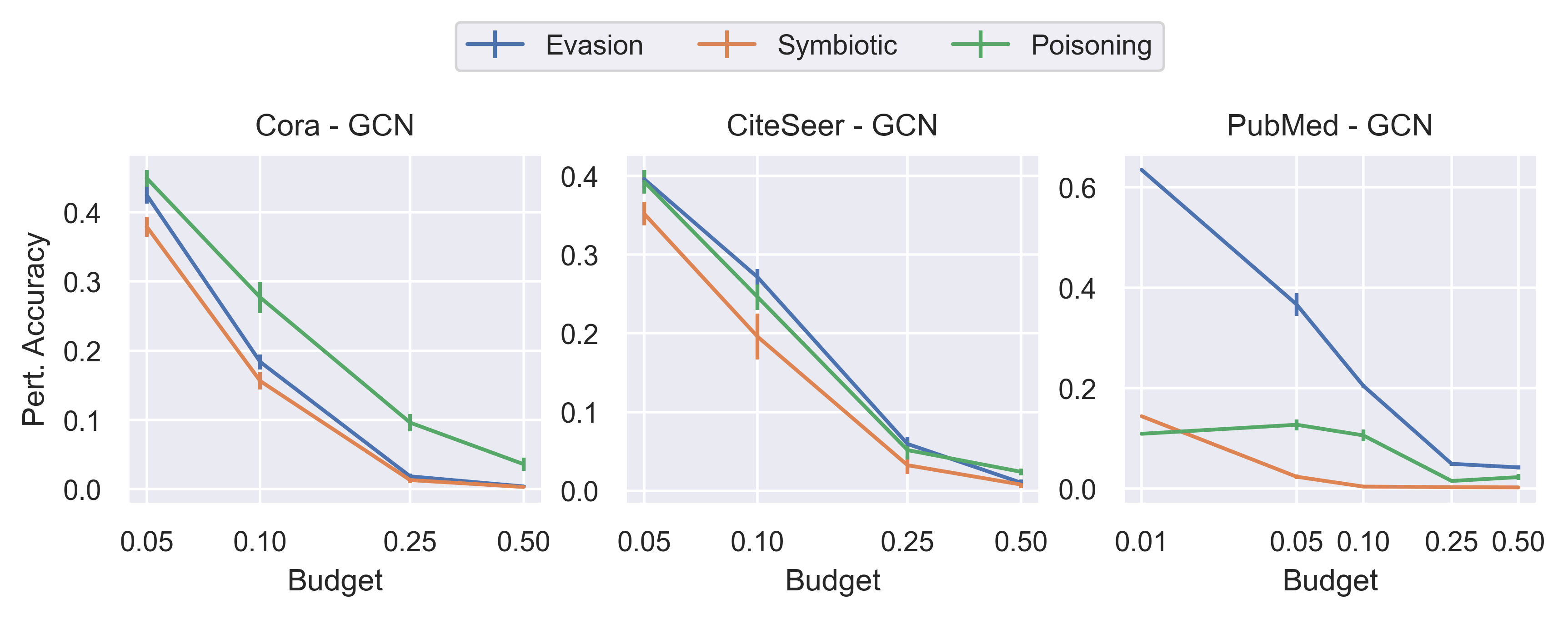}
    \caption{Perturbed accuracy of a GCN after the four attacks. The x-axis shows the global \textbf{budget} as a fraction of the number of edges.}
    \label{fig:budget}
\end{figure}

\subsection{Hyperparameters}

\textbf{Block size.} Figure \ref{fig:block_size} shows the results of the four attacks with varying block sizes, a fixed 5\% budget, and 125 attack iterations against a GCN. We observe that for very small block sizes, the attacks are less effective since the PR-BCD optimization can cover only a small part of the adjacency matrix. However, the marginal benefit of larger block sizes decreases once a large part of the adjacency matrix can be covered. 

\textbf{Budget.} Considering different fractions of the number of edges as the global budget, Figure \ref{fig:budget} indicates that all four attacks follow a similar trend with an increasing budget. Especially on PubMed where the share of labeled train nodes is the lowest, changing 5\% of the edges is enough to achieve close to zero accuracy under the symbiotic threat model. This highlights the potentially devastating effect of the joint attacks especially in large graphs with few labeled train nodes.

\section{Conclusion / Future Work}

In this work, we considered the symbiotic threat model of a combined test and train time adversary aiming to degrade the overall performance of a GNN on node classification tasks. We proposed two methods of obtaining perturbations, one a special case of the other, and demonstrated the potentially devastating impact such an attacker can have compared to a plain poisoning or evasion attacker. We now conclude by outlining potential lines of future work regarding the symbiotic threat model.

\textbf{Different Attacks as the Inner Evasion Step.} The joint attack we described is not limited to PR-BCD, and can be used with other evasion attacks, or attacks designed specifically for the symbiotic setting, to obtain perturbations within the poisoning attack.

\textbf{Inductive Tasks, Local Budgets, Targeted Attacks.} The combined threat model is also applicable to inductive tasks in which the test graph is different from the train graph. The poisoning half of the joint attack can be performed by targeting the validation nodes rather than the test nodes, and the evasion half is performed on the new test graph. Similarly, the attacks can also be applied with small changes under per-node local budgets, or target specific nodes. We plan to include further evaluations on these settings as our next step. 

\textbf{New Evasion-Aware Poisoning Attacks.} Finally, further novel poisoning attacks can also be developed which utilize the knowledge of a future evasion in different ways. 

\begin{ack}
This research was supported by the Helmholtz Association under the joint research school “Munich School for Data Science - MUDS“.
\end{ack}

\bibliographystyle{apalike}
\bibliography{references}

\newpage
\appendix

\begin{table}[h!]
    \caption{Perturbed accuracies (± standard error) of the joint and sequential attacks under the symbiotic threat model with a 5\% global budget. The -J suffix indicates the graph has been pre-processed with Jaccard purification \citep{wuAdversarialExamplesGraph2019}.}
    \label{tab:joint_sequential}
    \centering
    \begin{tabular}{lllll}
        \toprule
        \textbf{Model} & \textbf{Dataset} & \textbf{Clean} & \textbf{Sequential} & \textbf{Joint} \\
        \midrule
        \multirow[t]{3}{*}{GCN} 
        & CiteSeer & 0.68 ± 0.01 & 0.41 ± 0.01 & \textbf{0.38 ± 0.01} \\
         & CiteSeer-J & 0.68 ± 0.01 & 0.4 ± 0.01 & \textbf{0.38 ± 0.01} \\
         & Cora & 0.78 ± 0.01 & 0.37 ± 0.02 & \textbf{0.35 ± 0.01} \\
         & Cora-J & 0.74 ± 0.01 & \textbf{0.36 ± 0.01} & \textbf{0.36 ± 0.02} \\
          & PubMed & 0.78 ± 0.01 & 0.05 ± 0.01 & \textbf{0.03 ± 0.01} \\
         & PubMed-J & 0.77 ± 0.01 & 0.04 ± 0.01 & \textbf{0.02 ± 0.0} \\
        \midrule
        \multirow[t]{3}{*}{GAT} 
        & CiteSeer & 0.62 ± 0.02 & \textbf{0.3 ± 0.03} & 0.38 ± 0.02 \\
         & CiteSeer-J & 0.64 ± 0.01 & \textbf{0.3 ± 0.03} & 0.36 ± 0.02 \\
         & Cora & 0.69 ± 0.02 & \textbf{0.29 ± 0.02} & 0.32 ± 0.02 \\
         & Cora-J & 0.67 ± 0.01 & \textbf{0.28 ± 0.02} & 0.3 ± 0.03 \\
         & PubMed & 0.73 ± 0.01 & 0.24 ± 0.02 & \textbf{0.2 ± 0.03} \\
         & PubMed-J & 0.74 ± 0.01 & 0.27 ± 0.04 & \textbf{0.19 ± 0.02} \\
        \midrule
        \multirow[t]{3}{*}{APPNP} 
        & CiteSeer & 0.69 ± 0.01 & \textbf{0.47 ± 0.01} & 0.48 ± 0.01 \\
         & CiteSeer-J & 0.68 ± 0.01 & \textbf{0.45 ± 0.02} & \textbf{0.45 ± 0.02} \\
         & Cora & 0.82 ± 0.02 & 0.54 ± 0.02 & \textbf{0.51 ± 0.04} \\
         & Cora-J & 0.82 ± 0.01 & 0.57 ± 0.01 & \textbf{0.54 ± 0.01} \\
         & PubMed & 0.79 ± 0.0 & \textbf{0.09 ± 0.02} & \textbf{0.09 ± 0.01} \\
         & PubMed-J & 0.77 ± 0.01 & \textbf{0.1 ± 0.02} & 0.12 ± 0.02 \\
        \midrule
        \multirow[t]{3}{*}{GPRGNN} 
        & CiteSeer & 0.66 ± 0.01 & 0.34 ± 0.01 & \textbf{0.33 ± 0.01} \\
         & CiteSeer-J & 0.65 ± 0.01 & \textbf{0.35 ± 0.01} & \textbf{0.35 ± 0.01} \\
         & Cora & 0.82 ± 0.01 & 0.41 ± 0.01 & \textbf{0.4 ± 0.01} \\
         & Cora-J & 0.79 ± 0.01 & 0.42 ± 0.01 & \textbf{0.4 ± 0.01} \\
         & PubMed & 0.78 ± 0.01 & \textbf{0.08 ± 0.02} & 0.11 ± 0.03 \\
         & PubMed-J & 0.78 ± 0.01 & 0.16 ± 0.05 & \textbf{0.15 ± 0.04} \\
        \midrule
        \multirow[t]{3}{*}{RGCN} 
         & CiteSeer & 0.63 ± 0.01 & \textbf{0.47 ± 0.01} & \textbf{0.47 ± 0.01} \\
         & Cora & 0.74 ± 0.02 & 0.56 ± 0.01 & \textbf{0.52 ± 0.02} \\
         & PubMed & 0.77 ± 0.01 & 0.28 ± 0.04 & \textbf{0.15 ± 0.03} \\
        \bottomrule
    \end{tabular}
\end{table}

\begin{figure}[h]
    \centering
    \includegraphics[width=0.45\textwidth]{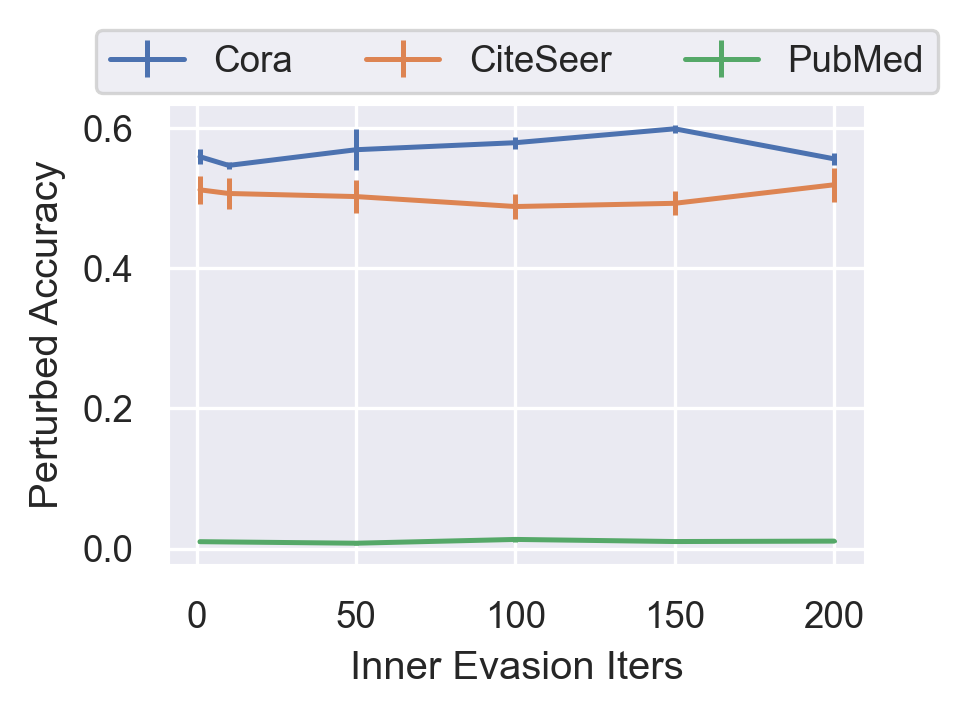}
    \caption{Perturbed accuracies and standard error after the joint attack on a GCN, with the x-axis indicating the number of iterations for the inner evasion part of the joint attack.}
    \label{fig:inner_evasion_epochs}
\end{figure}

\section{Comparing the Joint and Sequential Attacks} 

Extending on Table \ref{tab:main_results}, Table \ref{tab:joint_sequential} further compares the joint and sequential attacks of the symbiotic threat model. Although the joint attack is stronger than the sequential attack in a higher number of cases, they are often within one standard error of each other so it is difficult to argue for a statistically significant difference. This might indicate the difficulty of estimating the future evasion perturbations from within the poisoning attack due to the very large search space, and highlight the space for potential developments of stronger attacks. 

\section{Inner Evasion Iterations.} 

The number of iterations the evasion attack runs for within the joint attack is also configurable. Figure \ref{fig:inner_evasion_epochs} displays the accuracy of a GCN on our three benchmark graphs after the joint attack with different number of inner-evasion iterations. We observe, contrary to our expectations, that increasing the number of iterations for the inner-evasion attack has no significant influence on the perturbed accuracy. This perhaps highlights a potential point of improvement for future symbiotic attacks. 

\begin{table}[]
    \caption{Inductive}
    \label{tab:inductive_results}
    \centering
    \begin{tabular}{llllll}
        \toprule
        \textbf{Model} & \textbf{Dataset} & \textbf{Clean} & \textbf{Evasion} & \textbf{Poisoning} & \textbf{Symbiotic} \\
        \midrule
        \multirow[t]{2}{*}{GCN} 
         & CiteSeer & 0.67 ± 0.01 & 0.41 ± 0.01 & 0.62 ± 0.01 & \textbf{0.33 ± 0.01} \\
         & Cora & 0.75 ± 0.02 & 0.42 ± 0.01 & 0.68 ± 0.03 & \textbf{0.3 ± 0.01} \\
         \midrule
         \multirow[t]{3}{*}{GAT} 
         & CiteSeer & 0.68 ± 0.01 & \textbf{0.37 ± 0.01} & 0.64 ± 0.02 & 0.56 ± 0.02\\
         & Cora & 0.77 ± 0.01 & \textbf{0.21 ± 0.01} & 0.61 ± 0.04 & 0.35 ± 0.03 \\
        \midrule
        \multirow[t]{2}{*}{APPNP} 
         & CiteSeer & 0.71 ± 0.01 & 0.47 ± 0.01 & 0.66 ± 0.02 & \textbf{0.4 ± 0.01} \\
         & Cora & 0.82 ± 0.02 & 0.53 ± 0.02 & 0.78 ± 0.01 & \textbf{0.37 ± 0.01} \\
        \midrule
        \multirow[t]{2}{*}{GPRGNN} 
         & CiteSeer & 0.67 ± 0.01 & 0.37 ± 0.01 & 0.56 ± 0.01 & \textbf{0.34 ± 0.01} \\
         & Cora & 0.8 ± 0.02 & 0.44 ± 0.01 & 0.74 ± 0.01 & \textbf{0.35 ± 0.01} \\
        \bottomrule
    \end{tabular}
\end{table}

\end{document}

%% file: tables/main_results.tex
\begin{table}[t!]
    \centering
    \caption{Average (± standard error) perturbed accuracies for the evasion, poisoning, and symbiotic attacks with a 5\% budget.  The -J suffix indicates the graph has been pre-processed with Jaccard purification \citep{wuAdversarialExamplesGraph2019} and \textit{(ind.)} stands for inductive learning. The strongest (lowest accuracy) results for each setup are written in bold.}
    \label{tab:main_results}
    \begin{tabular}{llllll}
        \toprule
        \textbf{Model} & \textbf{Dataset} & \textbf{Clean} & \textbf{Evasion} & \textbf{Poisoning} & \textbf{Symbiotic} \\
        \midrule
        \multirow[t]{3}{*}{GCN} 
         & CiteSeer & 0.68 ± 0.01 & 0.41 ± 0.01 & 0.4 ± 0.01 & \textbf{0.38 ± 0.01} \\
         & CiteSeer (ind.) & 0.67 ± 0.01 & 0.41 ± 0.01 & 0.62 ± 0.01 & \textbf{0.33 ± 0.01} \\
         & CiteSeer-J & 0.68 ± 0.01 & 0.41 ± 0.01 & 0.41 ± 0.02 & \textbf{0.38 ± 0.01} \\
         & Cora & 0.78 ± 0.01 & 0.41 ± 0.01 & 0.46 ± 0.02 & \textbf{0.35 ± 0.01} \\
         & Cora (ind.) & 0.75 ± 0.02 & 0.42 ± 0.01 & 0.68 ± 0.03 & \textbf{0.3 ± 0.01} \\
         & Cora-J & 0.74 ± 0.01 & 0.39 ± 0.01 & 0.43 ± 0.02 & \textbf{0.36 ± 0.01} \\
         & PubMed & 0.78 ± 0.01 & 0.41 ± 0.01 & 0.12 ± 0.02 & \textbf{0.03 ± 0.01} \\
         & PubMed-J & 0.77 ± 0.01 & 0.41 ± 0.01 & 0.11 ± 0.01 &\textbf{0.02 ± 0.0} \\
        \midrule
        \multirow[t]{3}{*}{GAT} 
         & CiteSeer & 0.62 ± 0.02 & \textbf{0.27 ± 0.02} & 0.41 ± 0.02 & 0.3 ± 0.03 \\
         & CiteSeer (ind.) & 0.68 ± 0.01 & \textbf{0.37 ± 0.01} & 0.64 ± 0.02 & 0.56 ± 0.02\\
         & CiteSeer-J &  0.64 ± 0.01 & 0.32 ± 0.03 & 0.41 ± 0.03 & \textbf{0.3 ± 0.03} \\
         & Cora & 0.69 ± 0.02 & \textbf{0.22 ± 0.02} & 0.48 ± 0.03 & 0.29 ± 0.02 \\
         & Cora (ind.) & 0.77 ± 0.01 & \textbf{0.21 ± 0.01} & 0.61 ± 0.04 & 0.35 ± 0.03 \\
         & Cora-J & 0.67 ± 0.01 & \textbf{0.23 ± 0.02} & 0.45 ± 0.02 & 0.28 ± 0.02 \\
         & PubMed & 0.73 ± 0.01 & 0.38 ± 0.04 & 0.41 ± 0.01 & \textbf{0.2 ± 0.03} \\
         & PubMed-J & 0.74 ± 0.01 & 0.34 ± 0.04 & 0.38 ± 0.04 & \textbf{0.19 ± 0.02} \\
        \midrule
        \multirow[t]{3}{*}{APPNP} 
         & CiteSeer & 0.69 ± 0.01 & \textbf{0.45 ± 0.01} & 0.56 ± 0.01 & 0.47 ± 0.01 \\
         & CiteSeer (ind.) & 0.71 ± 0.01 & 0.47 ± 0.01 & 0.66 ± 0.02 & \textbf{0.4 ± 0.01} \\
         & CiteSeer-J & 0.68 ± 0.01 & \textbf{0.43 ± 0.01} & 0.52 ± 0.02 & 0.45 ± 0.02 \\
         & Cora & 0.82 ± 0.02 & \textbf{0.48 ± 0.03} & 0.64 ± 0.02 & 0.51 ± 0.04 \\
         & Cora (ind.) & 0.82 ± 0.02 & 0.53 ± 0.02 & 0.78 ± 0.01 & \textbf{0.37 ± 0.01} \\
         & Cora-J & 0.82 ± 0.01 & \textbf{0.5 ± 0.01} & 0.67 ± 0.01 & 0.54 ± 0.01 \\
         & PubMed & 0.79 ± 0.0 & 0.46 ± 0.01 & 0.21 ± 0.02 & \textbf{0.09 ± 0.01} \\
         & PubMed-J & 0.77 ± 0.01 & 0.45 ± 0.01 & 0.19 ± 0.03 & \textbf{0.1 ± 0.02} \\
        \midrule
        \multirow[t]{3}{*}{GPRGNN} 
         & CiteSeer & 0.66 ± 0.01 & 0.34 ± 0.01 & 0.44 ± 0.02 & \textbf{0.33 ± 0.01} \\
         & CiteSeer (ind.) & 0.67 ± 0.01 & 0.37 ± 0.01 & 0.56 ± 0.01 & \textbf{0.34 ± 0.01} \\
         & CiteSeer-J & 0.65 ± 0.01 & \textbf{0.35 ± 0.01} & 0.44 ± 0.01 & \textbf{0.35 ± 0.01} \\
         & Cora & 0.82 ± 0.01 & 0.46 ± 0.01 & 0.53 ± 0.01 & \textbf{0.4 ± 0.01} \\
         & Cora (ind.) & 0.8 ± 0.02 & 0.44 ± 0.01 & 0.74 ± 0.01 & \textbf{0.35 ± 0.01} \\
         & Cora-J &  0.79 ± 0.01 & 0.44 ± 0.01 & 0.54 ± 0.01 & \textbf{0.4 ± 0.01} \\
         & PubMed & 0.78 ± 0.01 & 0.42 ± 0.01 & 0.28 ± 0.03 & \textbf{0.08 ± 0.02} \\
         & PubMed-J & 0.78 ± 0.01 & 0.42 ± 0.01 & 0.38 ± 0.04 & \textbf{0.15 ± 0.04} \\
        \midrule
        \multirow[t]{3}{*}{RGCN} 
         & CiteSeer & 0.63 ± 0.01 & \textbf{0.39 ± 0.01} & 0.59 ± 0.02 & 0.47 ± 0.01 \\
         & Cora & 0.74 ± 0.02 & \textbf{0.44 ± 0.01} & 0.74 ± 0.01 & 0.52 ± 0.02 \\
         & PubMed & 0.77 ± 0.01 & 0.43 ± 0.01 & 0.42 ± 0.04 & \textbf{0.15 ± 0.03} \\
        \bottomrule
    \end{tabular}
\end{table}

%% file: main.bbl
\begin{thebibliography}{}

\bibitem[Bojchevski and G{\"u}nnemann, 2019]{bojchevskiAdversarialAttacksNode2019}
Bojchevski, A. and G{\"u}nnemann, S. (2019).
\newblock Adversarial {{Attacks}} on {{Node Embeddings}} via {{Graph Poisoning}}.

\bibitem[Chen et~al., 2020]{chen2020link}
Chen, J., Lin, X., Shi, Z., and Liu, Y. (2020).
\newblock Link prediction adversarial attack via iterative gradient attack.
\newblock {\em IEEE Transactions on Computational Social Systems}, 7(4):1081--1094.

\bibitem[Chien et~al., 2020]{chienAdaptiveUniversalGeneralized2020}
Chien, E., Peng, J., Li, P., and Milenkovic, O. (2020).
\newblock Adaptive {{Universal Generalized PageRank Graph Neural Network}}.
\newblock In {\em International {{Conference}} on {{Learning Representations}}}.

\bibitem[Dai et~al., 2018]{daiAdversarialAttackGraph2018}
Dai, H., Li, H., Tian, T., Huang, X., Wang, L., Zhu, J., and Song, L. (2018).
\newblock Adversarial {{Attack}} on {{Graph Structured Data}}.
\newblock In {\em Proceedings of the 35th {{International Conference}} on {{Machine Learning}}}, pages 1115--1124. {PMLR}.

\bibitem[Gasteiger et~al., 2018]{gasteigerPredictThenPropagate2018}
Gasteiger, J., Bojchevski, A., and G{\"u}nnemann, S. (2018).
\newblock Predict then {{Propagate}}: {{Graph Neural Networks}} meet {{Personalized PageRank}}.
\newblock In {\em International {{Conference}} on {{Learning Representations}}}.

\bibitem[Geisler et~al., 2021]{geislerRobustnessGraphNeural2021}
Geisler, S., Schmidt, T., {\c S}irin, H., Z{\"u}gner, D., Bojchevski, A., and G{\"u}nnemann, S. (2021).
\newblock Robustness of {{Graph Neural Networks}} at {{Scale}}.
\newblock In {\em Advances in {{Neural Information Processing Systems}}}, volume~34, pages 7637--7649. {Curran Associates, Inc.}

\bibitem[Gosch et~al., 2023a]{gosch_adversarial_training_2023}
Gosch, L., Geisler, S., Sturm, D., Charpentier, B., Z\"ugner, D., and G\"unnemann, S. (2023a).
\newblock Adversarial training for graph neural networks.
\newblock In {\em Neural Information Processing Systems, {NeurIPS}}.

\bibitem[Gosch et~al., 2023b]{gosch_revisiting_2023}
Gosch, L., Sturm, D., Geisler, S., and G{\"u}nnemann, S. (2023b).
\newblock Revisiting robustness in graph machine learning.
\newblock In {\em 11th International Conference on Learning Representations, {ICLR}}.

\bibitem[Guo et~al., 2021]{guo2021dockstream}
Guo, J., Janet, J.~P., Bauer, M.~R., Nittinger, E., Giblin, K.~A., Papadopoulos, K., Voronov, A., Patronov, A., Engkvist, O., and Margreitter, C. (2021).
\newblock Dockstream: a docking wrapper to enhance de novo molecular design.
\newblock {\em Journal of cheminformatics}, 13(1):1--21.

\bibitem[Hao et~al., 2020]{hao2020p}
Hao, J., Zhao, T., Li, J., Dong, X.~L., Faloutsos, C., Sun, Y., and Wang, W. (2020).
\newblock P-companion: A principled framework for diversified complementary product recommendation.
\newblock In {\em Proceedings of the 29th ACM International Conference on Information \& Knowledge Management}, pages 2517--2524.

\bibitem[Kipf and Welling, 2016]{kipfSemiSupervisedClassificationGraph2016}
Kipf, T.~N. and Welling, M. (2016).
\newblock Semi-{{Supervised Classification}} with {{Graph Convolutional Networks}}.
\newblock In {\em International {{Conference}} on {{Learning Representations}}}.

\bibitem[McCallum et~al., 2000]{mccallum2000automating}
McCallum, A.~K., Nigam, K., Rennie, J., and Seymore, K. (2000).
\newblock Automating the construction of internet portals with machine learning.
\newblock {\em Information Retrieval}, 3:127--163.

\bibitem[Mujkanovic et~al., 2022]{mujkanovicAreDefensesGraph2022}
Mujkanovic, F., Geisler, S., G{\"u}nnemann, S., and Bojchevski, A. (2022).
\newblock Are {{Defenses}} for {{Graph Neural Networks Robust}}?
\newblock {\em Advances in Neural Information Processing Systems}, 35:8954--8968.

\bibitem[Nesterov, 2012]{nesterovEfficiencyCoordinateDescent2012}
Nesterov, {\relax Yu}. (2012).
\newblock Efficiency of {{Coordinate Descent Methods}} on {{Huge-Scale Optimization Problems}}.
\newblock {\em SIAM Journal on Optimization}, 22(2):341--362.

\bibitem[Sen et~al., 2008]{sen2008collective}
Sen, P., Namata, G., Bilgic, M., Getoor, L., Galligher, B., and Eliassi-Rad, T. (2008).
\newblock Collective classification in network data.
\newblock {\em AI magazine}, 29(3):93--93.

\bibitem[Veli{\v c}kovi{\'c} et~al., 2018]{velickovicGraphAttentionNetworks2018a}
Veli{\v c}kovi{\'c}, P., Cucurull, G., Casanova, A., Romero, A., Li{\`o}, P., and Bengio, Y. (2018).
\newblock Graph {{Attention Networks}}.

\bibitem[Wang et~al., 2023]{wang2023revisiting}
Wang, X., Chang, H., Xie, B., Bian, T., Zhou, S., Wang, D., Zhang, Z., and Zhu, W. (2023).
\newblock Revisiting adversarial attacks on graph neural networks for graph classification.
\newblock {\em IEEE Transactions on Knowledge and Data Engineering}, pages 1--12.

\bibitem[Wu et~al., 2019]{wuAdversarialExamplesGraph2019}
Wu, H., Wang, C., Tyshetskiy, Y., Docherty, A., Lu, K., and Zhu, L. (2019).
\newblock Adversarial {{Examples}} on {{Graph Data}}: {{Deep Insights}} into {{Attack}} and {{Defense}}.

\bibitem[Xu et~al., 2019]{xuTopologyAttackDefense2019}
Xu, K., Chen, H., Liu, S., Chen, P.-Y., Weng, T.-W., Hong, M., and Lin, X. (2019).
\newblock Topology {{Attack}} and {{Defense}} for {{Graph Neural Networks}}: {{An Optimization Perspective}}.

\bibitem[Zhang et~al., 2019]{zhangDataPoisoningAttack2019}
Zhang, H., Zheng, T., Gao, J., Miao, C., Su, L., Li, Y., and Ren, K. (2019).
\newblock Data poisoning attack against knowledge graph embedding.
\newblock In {\em Proceedings of the 28th {{International Joint Conference}} on {{Artificial Intelligence}}}, {{IJCAI}}'19, pages 4853--4859, {Macao, China}. {AAAI Press}.

\bibitem[Zhu et~al., 2019]{zhuRobustGraphConvolutional2019}
Zhu, D., Zhang, Z., Cui, P., and Zhu, W. (2019).
\newblock Robust {{Graph Convolutional Networks Against Adversarial Attacks}}.
\newblock In {\em Proceedings of the 25th {{ACM SIGKDD International Conference}} on {{Knowledge Discovery}} \& {{Data Mining}}}, pages 1399--1407, {Anchorage AK USA}. {ACM}.

\bibitem[Z{\"u}gner et~al., 2018]{zugnerAdversarialAttacksNeural2018}
Z{\"u}gner, D., Akbarnejad, A., and G{\"u}nnemann, S. (2018).
\newblock Adversarial {{Attacks}} on {{Neural Networks}} for {{Graph Data}}.
\newblock In {\em Proceedings of the 24th {{ACM SIGKDD International Conference}} on {{Knowledge Discovery}} \& {{Data Mining}}}, {{KDD}} '18, pages 2847--2856, {New York, NY, USA}. {Association for Computing Machinery}.

\bibitem[Z{\"u}gner et~al., 2020]{zugnerAdversarialAttacksGraph2020}
Z{\"u}gner, D., Borchert, O., Akbarnejad, A., and G{\"u}nnemann, S. (2020).
\newblock Adversarial {{Attacks}} on {{Graph Neural Networks}}: {{Perturbations}} and their {{Patterns}}.
\newblock {\em ACM Transactions on Knowledge Discovery from Data}, 14(5):57:1--57:31.

\end{thebibliography}
